\documentclass[letterpaper, 10 pt, conference]{ieeeconf}  % Comment this line out if you need a4paper

\IEEEoverridecommandlockouts                              % This command is only needed if 
\usepackage{amsmath}
\usepackage{amssymb}
\usepackage{url}
\usepackage{makecell}
\usepackage{graphicx}
\usepackage{epstopdf}
\usepackage{cite}
\usepackage{epsfig} % for postscript graphics files
\usepackage[caption=false]{subfig}

\title{\LARGE \bf
NMPCB: A Lightweight and Safety-Critical Motion Control Framework for Ackermann Mobile Robot
}

\author{Longze Zheng and Qinghe Liu$^{\ast}$% <-this % stops a space
\thanks{
 All authors are with the School of Automotive Engineering at Harbin Institute of Technology, Weihai 264200, China. 23S030104@stu.hit.edu.cn, qingheliu@hitwh.edu.cn
}
\thanks{$\ast$ denotes the corresponding author}
}

\begin{document}

\maketitle
\thispagestyle{empty}
\pagestyle{empty}

%%%%%%%%%%%%%%%%%%%%%%%%%%%%%%%%%%%%%%%%%%%%%%%%%%%%%%%%%%%%%%%%%%%%%%%%%%%%%%%%
\begin{abstract}

In multi-obstacle environments, real-time performance and safety in robot motion control have long been challenging issues, as conventional methods often struggle to balance the two. In this paper, we propose a novel motion control framework composed of a \underline{N}eural network-based path planner and a \underline{M}odel \underline{P}redictive \underline{C}ontrol (MPC) controller based on control \underline{B}arrier function (NMPCB) . The planner predicts the next target point through a lightweight neural network and generates a reference trajectory for the controller. In the design of the controller, we introduce the dual problem of control barrier function (CBF) as the obstacle avoidance constraint, enabling it to ensure robot motion safety while significantly reducing computation time. The controller directly outputs control commands to the robot by tracking the reference trajectory. This framework achieves a balance between real-time performance and safety. We validate the feasibility of the framework through numerical simulations and real-world experiments.

\end{abstract}

%%%%%%%%%%%%%%%%%%%%%%%%%%%%%%%%%%%%%%%%%%%%%%%%%%%%%%%%%%%%%%%%%%%%%%%%%%%%%%%%
\section{\textsc{Introduction}}

In the design of robot motion planners and controllers, safety has always been the paramount requirement, considering the rapid development of robot systems\cite{ding2021epsilon}. The control method based on CBF has garnered significant attention for offering a simple yet safety-guaranteeing approach to robot motion control\cite{ames2019cbftheory}. 
In particular, the control method that integrates CBF and MPC has demonstrated superior performance in safety control by predicting future states \cite{zeng2021mpccbf}. However, this method often fails to obtain an optimal solution within a short computation time, especially for nonlinear kinematic systems where solution failure frequently occurs. Consequently, it also demands a high level of precision from the reference trajectory.

This paper presents a novel framework that integrates a neural network-based planner with an improved CBF-based MPC controller to ensure safety and real-time performance in robot motion control. The proposed framework serves both as a real-time local planner and as a controller.

\subsection{Related Work}
\subsubsection{Path Planner}
Sampling-based and search-based methods have been extensively studied, such as A*\cite{duchovn2014a*}, PRM\cite{bohlin2000prm}, and RRT\cite{radaelli2014rrt}. These methods are fundamental in the field of robotics for path planning and have been widely researched and implemented in various applications.
However, these methods often suffer from high computational costs and prolonged computation times, thus being only applicable to low-dimensional spaces.

Learning-based path planning methods are capable of rapidly generating high-quality paths, adapting to complex environments, and reducing reliance on prior knowledge\cite{wang2020neuralRRT}\cite{meng2022nrrrt}. For instance, neural A*\cite{yonetani2021na*}, which is based on deep learning, offers a more efficient approach to path generation.
However, since these methods can only plan paths offline, while the controllers can effectively track the paths, the prerequisite is the ability to adjust control inputs in real time.
Optimization-based methods can achieve real-time path planning but often exhibit slower planning speeds\cite{zhang2020OBCA}\cite{han2023rda}.

\begin{figure}[t]
\centering
 \includegraphics[width=0.5\textwidth]{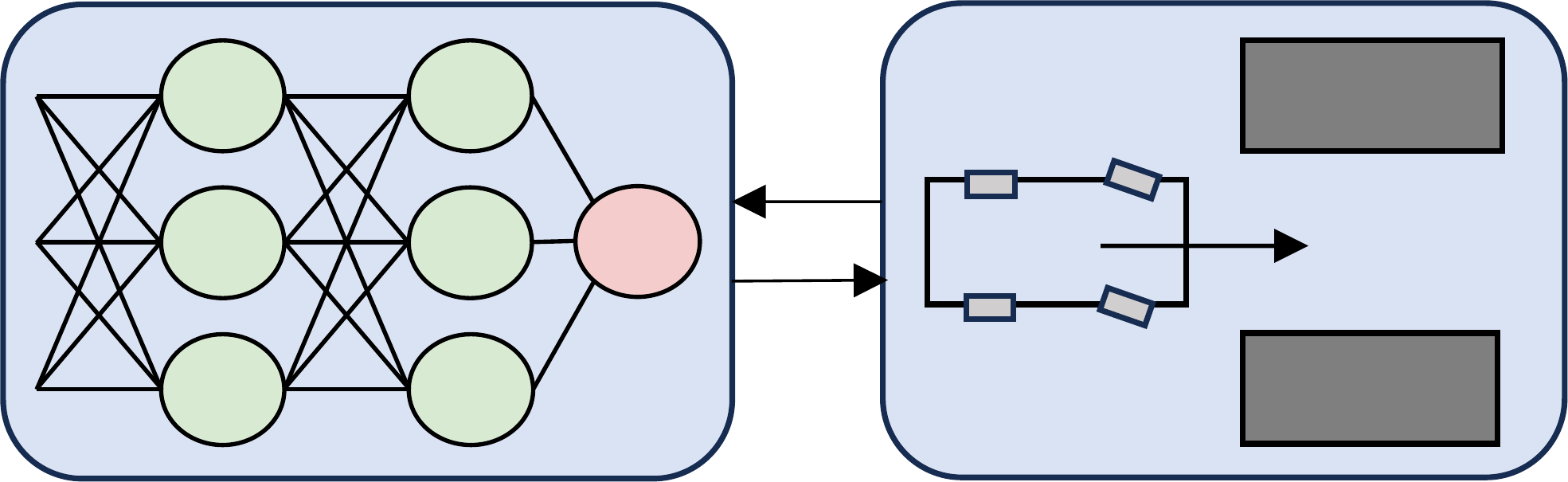}
  \caption{Schematic representation of the motion planning and control framework integrating a neural network-based planner and a CBF-based MPC controller for obstacle avoidance.}
  \label{fig1}
\end{figure}

\subsubsection{Optimization-Based Controller}
Collision-free optimal control means finding the maximum or minimum value of the cost function under constraints.
For instance, the TEB\cite{rosmann2017TEB} algorithm achieves motion control by solving a multi-objective optimization problem. MPC has demonstrated significant potential in collision-free tasks\cite{funke2016mpccolision}. 

The CBF serves as a safety boundary function for the control process, providing assurance of the robot's safety\cite{ames2019cbftheory}\cite{long2022safe}. CBF is particularly well-suited for collision avoidance constraints and has numerous applications in the fields of robotic manipulators \cite{dai2023arm} and vehicular systems \cite{seo2022vehicle}. In discrete-time systems, discrete-time control barrier function (DCBF) shows good performance in terms of safety constraints \cite{agrawal2017dcbf}. Research has been conducted on DCBF in the control and planning problems of multi-step optimization \cite{zeng2021dcbf_feasible}.

CBF-based MPC combines the advantages of both and has been increasingly used in recent years\cite{jian2023mpc_d_cbf}. Similarly, DCBF can also be combined with MPC to achieve better results in obstacle avoidance problems \cite{thirugnanam2022nmpc_dcbf}. The corresponding control method has been validated on both bipedal robots\cite{teng2021dcbf_leg} and vehicle systems\cite{ma2021dcbf_vehicle}. However, the above methods either oversimplify the control scenarios or have difficulty meeting the real-time requirements of control.

\subsection{Contributions}
The contributions of this paper are as follows:
\begin{itemize}
    \item We propose an encoder-decoder path planner that can predict a collision-free next target point using only historical trajectories and obstacle information, while significantly reducing the time overhead of the planner.
    % We propose an encoder-decoder path planning architecture. In the encoder, a reference path generated by Dubins curves is employed as the prior input for the kinematic feasibility domain. The decoder takes the sequence of historical path points and the rasterized environmental representation as inputs to predict the target point pose one second ahead.
    % \item Based on our planning architecture, we have improved the MPC-DCBF formulation to better accommodate obstacle avoidance tasks and ensure its real-time performance.
    \item We improve the MPC-DCBF formulation by constructing the dual form of obstacle avoidance constraints, which converts the implicit CBF constraints into explicit ones, thereby allowing it to better adapt to obstacle avoidance tasks and ensuring real-time performance.
    \item We conduct numerical simulations and real-world experiments for the proposed framework in this paper, and perform comparative experiments with several baseline schemes.
\end{itemize}
\subsection{Paper Structure}

The structure of this paper is organized as follows: in Section II, we present the preliminaries of MPC and CBF. In Section III, we introduce our proposed framework, along with the detailed design of the planner and controller. In Section IV, we describe our numerical simulations and real-world experiments. Finally, Section V concludes the paper.

% \begin{figure*}[htpb]
% \centering
%  \includegraphics[width=1\textwidth]{pic/nmpcb1.eps}
%   % \includegraphics[scale=0.3]{pic/nmpcb1.eps}
%   \caption{Inductance of oscillation winding on amorphous
%    magnetic core versus DC bias magnetic field}
%   \label{figurelabel}
% \end{figure*}
% --------------------------------------------------------------------------------------------------------------
\section{\textsc{Preliminaries}}

In this section, we will introduce MPC and the optimization formulation using DCBF as the collision avoidance constraint.
\subsection{Model Predictive Control}

The state transition equation of the robot is formulated as follows:
\begin{equation}
    \mathbf{x}_{t+1} = f(\mathbf{x}_t, \mathbf{u}_t),
\end{equation}
where \( \mathbf{x}_t \in \mathcal{X} \subset \mathbb{R}^n \) and \( \mathbf{u}_t \in \mathcal{U} \subset \mathbb{R}^m \) denote the state and control input of the robot at time step \( t \) respectively, and \( f \) is continuous. 

If the distance to obstacles is employed as a collision avoidance constraint, then the MPC control formulation at time $t$ is given as follows:
\begin{subequations}
\begin{align}
    \min_{\mathbf{u}_{t:t+N-1|t}} & \quad p(\mathbf{x}_{t+N|t}) + \sum_{k=0}^{N-1} q(\mathbf{x}_{t+k|t}, \mathbf{u}_{t+k|t}) \\
    \text{s.t.} & \quad \mathbf{x}_{t+k+1|t} = f(\mathbf{x}_{t+k|t}, \mathbf{u}_{t+k|t}),\notag\\ 
    & \quad \quad \text{for}\  k = 0, \ldots, N-1 \\
    & \quad \mathbf{x}_{t+k|t} \in \mathcal{X}, \, \mathbf{u}_{t+k|t} \in \mathcal{U},\notag\\
    & \quad \quad \text{for} \, k = 0, \ldots, N-1  \\
    & \quad \mathbf{x}_{t|t} = \mathbf{x}_t,\  \mathbf{x}_{t+N|t} \in \mathcal{X}_f\\
    & \quad g(\mathbf{x}_{t+k+1|t}) \geq 0, \, k = 0, \ldots, N-1. 
\end{align}
\end{subequations}
Here, the functions $p(\cdot)$ and $q(\cdot)$ in (2a) represent the terminal cost and the cost associated with tracking the reference trajectory, respectively. Constraint (2d) indicates that the initial state of the robot is $\mathbf{x}_t$ and the subsequent states are predicted via the system dynamics given in (2b), and constraint (2d) also specifies the terminal constraint.  
Both state and control input constraints are provided by (2c). The collision avoidance constraint is given by (2e), which can be defined using various Euclidean norms.
$N$ denotes the prediction horizons.

We can also regard the MPC as a path planning module. The path planning problem can be formulated as planning a collision-free trajectory for the robot from the initial state $\mathbf{x}_0$ to the target state $\mathbf{x}_{\text{goal}}$. If we set the length of the prediction horizons $N$ in the optimization model (2) to $1$ and solve for the optimal state $\mathbf{x}$, this becomes a standard path planning problem. We will employ an encoder-decoder model to address the path planning problem, the specifics of which will be discussed in Section III-A.

\subsection{Control Barrier Function}
If the dynamical system (1) is safe with respect to the set $S \subseteq \mathcal{X}$, then any trajectory initiated from within $S$ will remain within $S$. The set $S$ is defined as the 0-superlevel set of a continuous function $h: \mathcal{X} \rightarrow \mathbb{R}$ as:
\begin{equation}
    \mathcal{S} := \{ \mathbf{x} \in \mathcal{X} \subset \mathbb{R}^n : h(\mathbf{x}) \geq 0 \}.
\end{equation}
We refer to $S$ as the safety set, which encompasses all regions without obstacles.  $h$ is defined as a DCBF, if satisfied
\begin{equation}
\begin{aligned}
     & h(f(\mathbf{x}, \mathbf{u})) \geq \gamma(\mathbf{x}) h(\mathbf{x}),\\ 
     & 0 \leq \gamma(\mathbf{x}) < 1, \forall \mathbf{x} \in \mathcal{S} ,\  \exists \mathbf{u} \in \mathcal{U}.
\end{aligned}
\end{equation}
Let $\gamma _k := \gamma(\mathbf{x}_k)$. Satisfying (4) implies $h(f(\mathbf{x}, \mathbf{u})) \geq \gamma_k h(\mathbf{x})$, that is, the lower bound of the DCBF decreases exponentially with the decay rate $\gamma _k$ \cite{zeng2021dcbf_feasible}. We denote $\mathcal{K}(\mathbf{x})$ as
\begin{equation}
    \mathcal{K}(\mathbf{x}) := \{ \mathbf{u} \in \mathcal{U} : h(f(\mathbf{x}, \mathbf{u})) - \gamma(\mathbf{x})h(\mathbf{x}) \geq 0 \}.
\end{equation}
Then, if the initial state is within the safety set $S$ and $\mathbf{u}_k \in \mathcal{K}(\mathbf{x}_k)$, all subsequent states will also remain within the safety set $S$,  which implies that the resulting trajectory is safe.

The formulation proposed later in \cite{zeng2021dcbf_feasible} introduces a slack variable $\omega$ to balance feasibility and safety as follows:
\begin{equation}
    h(f(\mathbf{x}, \mathbf{u})) \geq \omega(\mathbf{x}) \gamma(\mathbf{x}) h(\mathbf{x}), \quad 0 \leq \gamma(\mathbf{x}) < 1.
\end{equation}
The MPC control formulation based on DCBF constraint (MPC-DCBF) at time $t$ is presented as follows:
\begin{subequations}
\begin{align}
    \min_{U,\Omega} & \quad p(\mathbf{x}_{t+N|t}) + \sum_{k=0}^{N-1} q(\mathbf{x}_{t+k|t}, \mathbf{u}_{t+k|t}) + \psi(\omega _k)  \\
    \text{s.t.} & \quad \mathbf{x}_{t+k+1|t} = f(\mathbf{x}_{t+k|t}, \mathbf{u}_{t+k|t}),\notag\\ 
    & \quad \quad \text{for}\  k = 0, \ldots, N-1 \\
    & \quad \mathbf{x}_{t+k|t} \in \mathcal{X}, \, \mathbf{u}_{t+k|t} \in \mathcal{U},\notag\\
    & \quad \quad \text{for} \, k = 0, \ldots, N-1  \\
    & \quad  h(\mathbf{x}_{t+k+1|t}) \geq \omega_k \gamma_k h(\mathbf{x_{t+k|t}}),\  \omega_k \geq 0 , \notag\\ 
    & \quad \quad \text{for} \ k = 0, \ldots, N_{\text{CBF}}-1,\\
    & \quad \mathbf{x}_{t|t} = \mathbf{x}_t,\  \mathbf{x}_{t+N|t} \in \mathcal{X}_f.
\end{align}
\end{subequations}

\begin{figure*}[thbp]
\centering
 \includegraphics[width=0.95\textwidth]{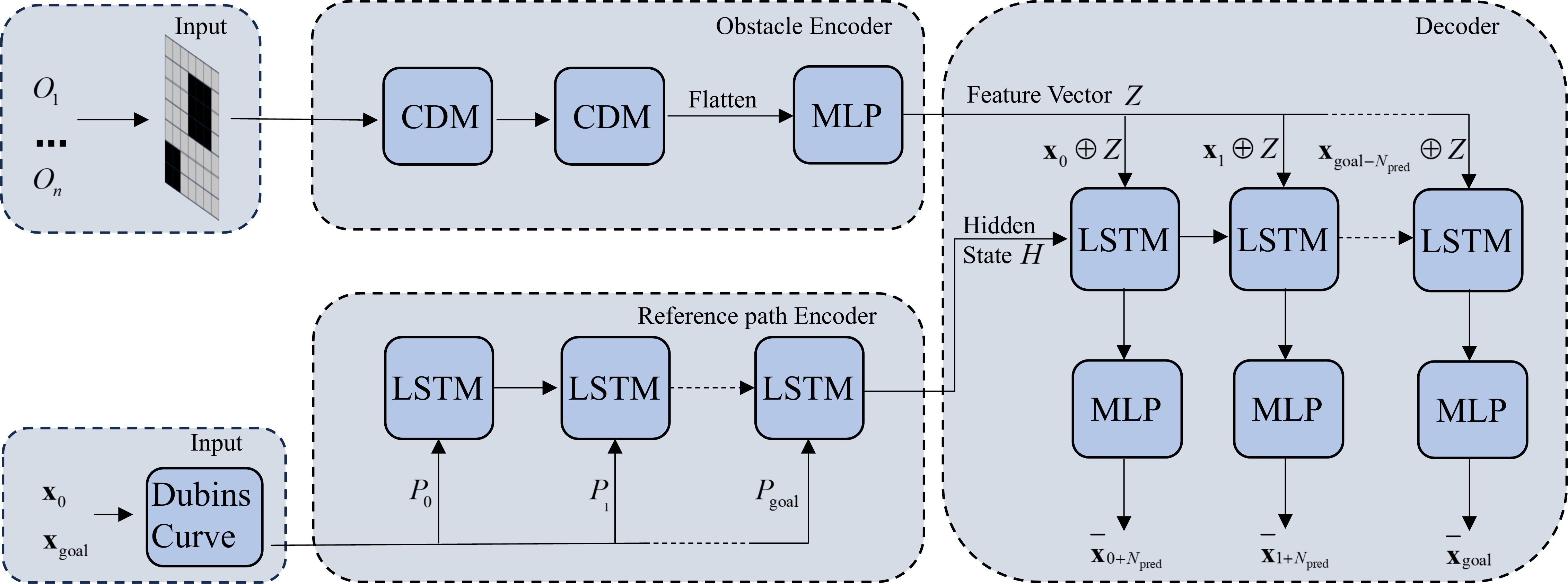}
  \caption{Illustration of the Neural Dubins Model.}
  \label{fig2}
\end{figure*}

The optimization formulation (7) is fundamentally similar to the optimization formulation (2). In formulation (7), $U = [\mathbf{u}^T_{t|t},\ldots,\mathbf{u}^T_{t+N-1|t}]$ and $\Omega = [\omega_0,\ldots,\omega_{N_{\text{CBF}-1}}]$ are the control input and relaxation variables, respectively. $\psi$ is the penalty function for the relaxation variable. $N_{\text{CBF}}$ denotes the safety horizon, which must satisfy $N_{\text{CBF}} \leq N$.
% --------------------------------------------------------------------------------------------------------------

\subsection{Obstacle Avoidance Problem}
To simplify the formulation of the obstacle avoidance problem, we introduce the following assumptions. First, the robot can be abstracted into a point mass model. Second, the geometry of any obstacle can be over-approximated with a union of convex polytopes, which is defined as a bounded polyhedron. In the $l$-dimensional space, the $i$-th obstacle can be described as:
\begin{equation}
    \mathcal{O}_i := \{ y \in \mathbb{R}^l : A^{\mathcal{O}_i} y \leq b^{\mathcal{O}_i} \} ,
\end{equation}
where $b^{\mathcal{O}_i} \in R^{s^{\mathcal{O}_i}}$. $s^{\mathcal{O}_i}$ represents the number of facets of polytopic sets for the $i$-th obstacle.

Let the state of the robot be $\mathbf{x} \in \mathbb{R}^n$ with its discrete-time dynamics as defined in (1) , and let $\mathcal{R}(\mathbf{x})$ represent the center of the robot. For $\mathbf{x} \in \mathcal{X}$, the computation of the squared minimum distance between the $i$-th obstacle $\mathcal{O}_i$ and the center $\mathcal{R}(\mathbf{x})$ can be denoted by the function $h_i(\mathbf{x})$ formula:
\begin{subequations}
\begin{align}
h_i(\mathbf{x}) = & \min_{y^{\mathcal{O}_i} \in \mathbb{R}^{l},\mathbf{x} \in \mathbb{R}^n} \|y^{\mathcal{O}_i} - \mathcal{R}(\mathbf{x})\|_2^2 \\
& \quad \quad\text{s.t.} \quad\quad A^{\mathcal{O}_i} y^{\mathcal{O}_i} \leq b^{\mathcal{O}_i}.
\end{align}
\end{subequations}

It can be noted that (9) formulates a quadratic programming (QP) problem, thus representing a convex optimization problem. To ensure safety throughout the motion process, we apply the same DCBF constraint to each obstacle in relation to the robot. Thus, the safety set is defined as:
\begin{equation}
    \mathcal{S}_i := \{ \mathbf{x} \in \mathbb{R}^n : h_i(\mathbf{x}) > 0 \}^c, \text{for} \ i = 1,\ldots,N_o,
\end{equation}
where $(\cdot)^c$ denotes the closure of a set, and $N_O$ denotes the number of obstacles. Enforcing the DCBF constraint for each $h_i$ ensures that the state remains in $\mathcal{S}:= \bigcap_{i=1}^{N_o} S_i$.

\section{\textsc{NMPCB Motion Control Framework}}
In this section, we introduce the NMPCB framework, which consists mainly of two components: 
the neural network-based path planner and the DCBF-based MPC controller.
\subsection{Neural Dubins Model}

\begin{figure}[ht]
\centering
 \includegraphics[width=0.45\textwidth]{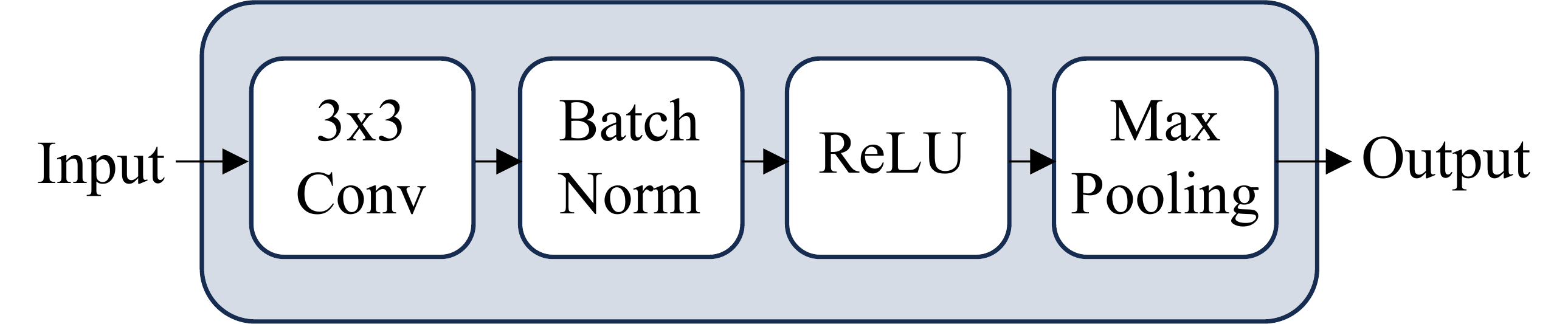}
  \caption{Illustration of the CDM.}
  \label{fig3}
\end{figure}

In this section, we describe our encoder-decoder model formulation. As shown in Fig. \ref{fig2} and Fig. \ref{fig3}, our model mainly consists of three components: the reference path encoder, the obstacle encoder, and the decoder.

The reference path encoder receives information from the start point $\mathbf{x}_0 = [x_0, y_0, \theta _0]$ and the goal point $\mathbf{x}_{\text{goal}} = [x_{\text{goal}}, y_{\text{goal}}, \theta _{\text{goal}}]$ as input, where $x$ and $y$ represent the position of the robot, and $\theta$ denotes the heading angle of the robot. Within the model, a Dubins curve\cite{shkel2001dubins} is first generated from $\mathbf{x}_0$ to $\mathbf{x}_{\text{goal}}$, and the coordinates of each point $P_k = [x_k,y_k,\theta _k]$ on the curve are sequentially input into a single-layer LSTM \cite{hochreiter1997lstm} module to obtain the hidden state $H$. LSTM networks exhibit superior adaptability to motion continuity and enhanced robustness to noisy or irregular path data when processing path sequences. 

The obstacle encoder receives obstacle information as input. Each obstacle can be represented as $O_i = [x_i,y_i,r_i]$, where $x$ and $y$ represent the center, and $r$ denotes the obstacle radius. The model maps the obstacle information onto a $50\times 50$ map matrix, where matrix values are set to $1$ at grid points with obstacles and $0$ at obstacle-free points, thus, the dimension of each map matrix is $(50,50,1)$. This map matrix is then fed into a hierarchical feature extraction architecture comprising two convolutional downsampling modules (CDM) and a fully connected compression module, which ultimately outputs a low-dimensional feature vector $Z$. 
Each CDM comprises individual two-dimensional convolutional layers with a kernel size of $3\times3$. Each convolutional layer is followed by batch normalization (BN), ReLU activation, and a $2\times2$ max pooling layer for downsampling.

Subsequently, the hidden state $H$ of the reference path encoder, the feature vector $Z$ of the obstacle encoder, the endpoint $\mathbf{x}_{\text{goal}}$, and all path points from the start point to the current point  $\mathbf{x} _{0:t}$ are input into the decoder. It then outputs the future states $\hat{\mathbf{{x}}} _{t+N_{\text{pred}}}$ within a specified time horizon $N_{\text{pred}}$,
\begin{equation}
    \mathcal{D}(H, Z, \mathbf{x}_{\text{goal}}, \mathbf{x}_{0:t}) \rightarrow \hat{\mathbf{{x}}} _{t+N_{\text{pred}}}.
\end{equation}
The decoder network $\mathcal{D}$ is composed of a single-layer LSTM network followed by a fully connected layer. We use a mean squared error (MSE) loss between the predicted state $\hat{\mathbf{{x}}} _{t+N_{\text{pred}}}$ and the label next state $\mathbf{{x}}_{t+N_{\text{pred}}}$ during training. The loss function $\mathcal{L}$ is defined as follows:
\begin{equation}
    \mathcal{L}(\hat{\mathbf{x}}, \mathbf{x}) = \frac{1}{N} \sum_{k=0}^{N}\left\| \hat{\mathbf{x}}_{k+N_{\text{pred}}} - \mathbf{x}_{k+N_{\text{pred}}} \right\|^2_2.
\end{equation}
Here, $N$ denotes the total number of planning steps.

After deriving the predicted state $\hat{\mathbf{{x}}} _{t+N_{\text{pred}}}$, a Dubins curve is constructed from the current point $\mathbf{x}_t$ to $\hat{\mathbf{{x}}} _{t+N_{\text{pred}}}$ to serve as a reference trajectory that is transmitted to the control module. We refer to the above-mentioned neural network-based path planner as the Neural Dubins Model.

\subsection{Dual DCBF Constraint}
We note that in the DCBF constraint (4), the computation of $h(f(\mathbf{x},\mathbf{u}))$ is required, which leads to the presence of non-differentiable implicit constraints in the optimization formulation and can consume significant computational time. Next, we introduce the dual DCBF constraint to address this issue.

For any convex optimization problem, a dual problem exists. The dual form of problem (9) is given by:
\begin{subequations}
\begin{align}
g_i(\mathbf{x}) = \max_{\lambda^{\mathcal{O}_i}}& \quad(A^{\mathcal{O}_i}\mathcal{R}(\mathbf{x})-b^{\mathcal{O}_i})^T\lambda^{\mathcal{O}_i}\\
\text{s.t.}  &\quad\| \lambda^{\mathcal{O}_i} A^{\mathcal{O}_i} \|_2 \leq 1,\\
&\quad \lambda^{\mathcal{O}_i}\geq 0 .
\end{align}
\end{subequations}
Here $(A^{\mathcal{O}_i})^{T}\lambda^{\mathcal{O}_i} $ is the normal vector of the separating hyperplane.

According to the Weak Duality Theorem, for all optimization problems, it holds that $g_i(\mathbf{x}) \leq h_i(\mathbf{x})$ for their respective dual problems. Since (9) is a convex optimization with linear constraint and has a well-defined optimum solution in \( \mathbb{R}^+ \), the Strong Duality Theorem\cite{boyd2004convex} also holds, which states that
\begin{equation}
   g_i(\mathbf{x}) = h_i(\mathbf{x}).
\end{equation}

In accordance with the  Strong Duality Theorem (14), we can substitute \( g_i(\mathbf{x}) \) for the computation of \( h_i(\mathbf{x}) \), thereby circumventing the implicit dependence of \( h_i(\mathbf{x}) \) on \( \mathbf{x} \).
We assume \( g_i(\mathbf{x}, \lambda^{\mathcal{O}_i}) \) represents the cost associated with any feasible solution \( \lambda^{\mathcal{O}_i} \) of (13). Since (13) is a maximization problem, we can derive the following inequality relationship:
\begin{equation}
   \bar{g}_i(\mathbf{x}, \lambda^{\mathcal{O}_i}) := (A^{\mathcal{O}_i}\mathcal{R}(\mathbf{x})-b^{\mathcal{O}_i})^T\lambda^{\mathcal{O}_i} \leq g_i(\mathbf{x}) = h_i(\mathbf{x}).
\end{equation}
Then, we can transform the DCBF constraint into a stronger form:
\begin{equation}
   (A^{\mathcal{O}_i}\mathcal{R}(f(\mathbf{x},\mathbf{u}))-b^{\mathcal{O}_i})^T\lambda^{\mathcal{O}_i} \geq \gamma _kh_i(\mathbf{x}).
\end{equation}
This represents a stronger DCBF constraint that satisfies the DCBF constraint (4). The substitution of the DCBF constraint with (16) requires satisfying (13b) and (13c), as follows:
\begin{subequations}
\begin{align}
&\quad(A^{\mathcal{O}_i}\mathcal{R}(f(\mathbf{x},\mathbf{u}))-b^{\mathcal{O}_i})^T\lambda^{\mathcal{O}_i} \geq \gamma _kh_i(\mathbf{x}),\\
&\quad\| \lambda^{\mathcal{O}_i} A^{\mathcal{O}_i} \|_2 \leq 1,\ \lambda^{\mathcal{O}_i}\geq 0.
\end{align}
\end{subequations}
According to the Strong Duality Theorem (14), $\exists\ \lambda^{\mathcal{O}_i *}$ satisfying (13b) and (13c) such that for all \(x \in \mathcal{X}\),
\begin{equation}
   \bar{g}_i(\mathbf{x}, \lambda^{\mathcal{O}_i *}) = (A^{\mathcal{O}_i}\mathcal{R}(\mathbf{x})-b^{\mathcal{O}_i})^T\lambda^{\mathcal{O}_i *} = h_i(\mathbf{x}).
\end{equation}
This implies that if and only if \( (\mathbf{u},\lambda^{\mathcal{O}_i *}) \) satisfy (16), for all fixed \( \mathbf{x} \in \mathcal{X} \), the input \( \mathbf{u} \) satisfies the DCBF constraint (4) with \( h_i \) implicitly defined, which also implies that the set of feasible inputs does not diminish at any \( \mathbf{x} \in \mathcal{X} \).

Similarly, we can apply the aforementioned method to optimization problem (9) to establish stronger DCBF constraint. Let \( \mathbf{y}^{\mathcal{O}_i} \) be any feasible solution to (9). Since (9) is a minimization problem, we can derive the following inequality relationship:
\begin{equation}
   \bar{h}_i(\mathbf{x}, y^{\mathcal{O}_i}) := \| y^{\mathcal{O}_i} - \mathcal{R}(\mathbf{x}) \|_2^2 \geq h_i(\mathbf{x}).
\end{equation}
Then we can enforce the stronger DCBF constrain:
\begin{equation}
  \quad(A^{\mathcal{O}_i}\mathcal{R}(f(\mathbf{x},\mathbf{u}))-b^{\mathcal{O}_i})^T\lambda^{\mathcal{O}_i} \geq \gamma _k\| y^{\mathcal{O}_i} - \mathcal{R}(\mathbf{x}) \|_2^2.
\end{equation}
Comparing with (4), we can observe the transformation of the DCBF constraint:
\begin{align}
  h_i(f(\mathbf{x},\mathbf{u})) &\geq (A^{\mathcal{O}_i}\mathcal{R}(f(\mathbf{x},\mathbf{u}))-b^{\mathcal{O}_i})^T\lambda^{\mathcal{O}_i} \notag\\ 
  &\geq \gamma _k\| y^{\mathcal{O}_i} - \mathcal{R}(\mathbf{x}) \|_2^2 \geq
  \gamma _k h_i(\mathbf{x}).
\end{align}

Integrating the analysis presented above, we define the dual DCBF constraint as follows:
\begin{subequations}
\begin{align}
&(A^{\mathcal{O}_i}\mathcal{R}(f(\mathbf{x},\mathbf{u}))-b^{\mathcal{O}_i})^T\lambda^{\mathcal{O}_i} \geq \gamma _k\| y^{\mathcal{O}_i} - \mathcal{R}(\mathbf{x}) \|_2^2,\\
&\| \lambda^{\mathcal{O}_i} A^{\mathcal{O}_i} \|_2 \leq 1,\ \lambda^{\mathcal{O}_i}\geq 0,\\
&A^{\mathcal{O}_i} y^{\mathcal{O}_i} \leq b^{\mathcal{O}_i}.
\end{align}
\end{subequations}
Introducing a slack variable \( \omega \) has no effect on the aforementioned formulation.

\subsection{Optimization Formulation}
We apply the dual DCBF constraint to the MPC algorithm as collision avoidance constraints to construct a multi-step optimization model, thereby providing enhanced assurance of safety in motion control.
The optimization formulation for MPC-DUAL-DCBF (MDD) at time $t$ is presented as follows:
\begin{subequations}
\begin{align}
        \min_{U,\Omega} & \quad p(\mathbf{x}_{t+N|t}) + \sum_{k=0}^{N-1} q(\mathbf{x}_{t+k|t}, \mathbf{u}_{t+k|t}) + \psi(\omega _k)  \\
    \text{s.t.} & \quad \mathbf{x}_{t|t} = \mathbf{x}_t, \ y_0^{\mathcal{O}_i}=y_t^{\mathcal{O}_i *},\\
    & \quad \mathbf{x}_{t+k|t} \in \mathcal{X}, \, \mathbf{u}_{t+k|t} \in \mathcal{U},\notag\\
    & \quad \quad \text{for} \, k = 0, \ldots, N-1  \\
    & \quad \mathbf{x}_{t+k+1|t} = f(\mathbf{x}_{t+k|t}, \mathbf{u}_{t+k|t}),\notag\\ 
    & \quad \quad \text{for}\  k = 0, \ldots, N-1 \\
    & \quad (A^{\mathcal{O}_i}\mathcal{R}(\mathbf{x}_{t+k+1|t})-b^{\mathcal{O}_i})^T\lambda^{\mathcal{O}_i}_{k+1} \notag\\
    & \quad \geq \omega _k \gamma _k\| y^{\mathcal{O}_i}_{k} - \mathcal{R}(\mathbf{x}_{t+k|t}) \|_2^2,\notag\\
    & \quad \quad \text{for}\  k = 0, \ldots, N_{\text{CBF}}-1 \\
    & \quad \| \lambda^{\mathcal{O}_i}_{k} A^{\mathcal{O}_i} \|_2 \leq 1, 
    \ A^{\mathcal{O}_i} y^{\mathcal{O}_i}_k \leq b^{\mathcal{O}_i},\notag \\
    & \quad \quad \text{for}\  k = 0, \ldots, N_{\text{CBF}}-1 \\
    & \quad \lambda^{\mathcal{O}_i}_k \geq 0,\ \omega_k \geq 0 .\notag \\
    & \quad \quad \text{for}\  k = 0, \ldots, N_{\text{CBF}}-1 
\end{align}
\end{subequations}
In the formulation, $y_t^{\mathcal{O}_i *}$ represents the optimal solution obtained by precomputing the minimum distance at time \( t \) through (9). Constraint (23e) represents the dual DCBF constraint, while  (23f)-(23g) denote the feasibility conditions of the dual DCBF constraint. The optimization formulation (23) illustrates only the dual DCBF constraint between obstacle \( \mathcal{O}_i \) and the robot \( \mathcal{R} \), however, during the motion control process, the corresponding dual DCBF constraint are applied to each pair of obstacles and the robot.

If the same dual DCBF constraint are imposed at each time step within a multi-step optimization model, it would result in a substantial computational burden. To reduce this complexity, we roll out the time \( k \) and replace the RHS of each DCBF constraint with \( h_i(\mathbf{x}_t) \), as follows:
\begin{equation}
  \quad(A^{\mathcal{O}_i}\mathcal{R}(\mathbf{x}_{t+k+1})-b^{\mathcal{O}_i})^T\lambda^{\mathcal{O}_i}_{t+k+1} \geq 
  \omega_k(\prod_{j=0}^{k} \gamma_j)h_i(\mathbf{x}_t).
\end{equation}
Replacing (23e) with (24) can accelerate the computation. This modification affects neither the system feasibility nor its safety.
\subsection{Framework Overview}
The framework consists of a planner and a controller. The planner utilizes Neural Dubins Model to determine the next target point and generates a Dubins curve from the current point to the predicted target point, which serves as a reference trajectory and is passed to the control module. The control module solves optimization problem (23) to output the optimal control commands, ensuring that the robot moves towards the goal without collisions. 

We focus on the motion control problem of mobile robots, hence adopting the bicycle model as the kinematic model. The state vector of the robot \( \mathbf{x} \) is defined as \( [x, y, v, \theta] \), where \( x \) and \( y \) represent the coordinates of the rear axle center, \( v \) is the velocity, and \( \theta \) is the yaw angle. The control vector \( \mathbf{u} \) is defined as \( [a, \delta] \), where \( a \) is the acceleration and \( \delta \) is the steering angle of the front wheel. The kinematic equations are given by:
\begin{subequations}
\begin{align}
    & x_{k+1} =x_k+ v_k\cos{(\theta_k)}\Delta t,\\
    & y_{k+1} =y_k+ v_k\sin{(\theta_k)}\Delta t,\\
    & v_{k+1} =v_k+ a_k\Delta t,\\
    & \theta_{k+1} =\theta_k+ \frac{v_k\tan{(\delta_k)}}{L}\Delta t,
\end{align}
\end{subequations}
where $\Delta t$ denotes the time step and $L$ denotes the wheelbase.

The cost function (23a) is composed of terminal cost, stage cost, and slack function cost, respectively as
\begin{subequations}
\begin{align}
    p(\mathbf{x}_{t+N|t}) &= \|\mathbf{x}_{t+N|t} - \bar{\mathbf{x}}_{t+N|t}\|^2, \\
q(\mathbf{x}_{t+k|t}, \mathbf{u}_{t+k|t}) &= \|\mathbf{x}_{t+k|t} - \bar{\mathbf{x}}_{t+k|t}\|^2 + \|\mathbf{u}_{t+k|t}\|^2 \nonumber \\
&\quad + \|\mathbf{u}_{t+k|t} - \mathbf{u}_{t+k-1|t}\|^2, \\
\psi(\omega_k) &= p_\omega (\omega_k - 1)^2,
\end{align}
\end{subequations}
where \( \bar{\mathbf{x}} \) represents the reference trajectory. The stage cost consists of the tracking cost of the reference trajectory, the control effort cost, and the control smoothness cost.

For the point mass model of the robot, we can ensure effective obstacle avoidance by setting a safety distance $D_{\text{safe}}$, which also alleviates the issue of decreased solution speed due to the incorporation of distant obstacles in the optimization problem.

For optimization problem (23), the geometric representation of the robot and obstacles can be altered to accommodate various scenarios. In scenarios where high solution speed is required and accurate shape information of obstacles can be obtained, an optimization formulation corresponding to a point mass robot model and convex polygon obstacles can be utilized (MDD-I). 
% In scenarios where accurate obstacle shape information cannot be obtained, an optimization formulation corresponding to a convex polygonal robot model and circular obstacles can be employed (MDD-II). Its optimization problem formulation aligns with (23), except that the dual DCBF constraints are constructed using the robot's dual variables.
In scenarios where solution speed is not a critical factor and navigating through tight spaces is challenging, an optimization model that corresponds to a convex polygonal robot model and convex polygonal obstacles can be applied (MDD-II, \cite{thirugnanam2022nmpc_dcbf}).

\section{\textsc{Experiments and Results}}

In this section, we perform numerical simulations of the NMPCB framework and conduct repeated experiments across diverse scenarios to verify that the framework can significantly improve the solution speed of the motion control problem while ensuring robot safety.
Furthermore, we have conducted real-world experiments with the proposed framework to validate its performance in actual environments.
\begin{figure*}[!t]
	\centering
	\subfloat[RDA Planner + MPC-DCBF]{\includegraphics[width=1.75in]{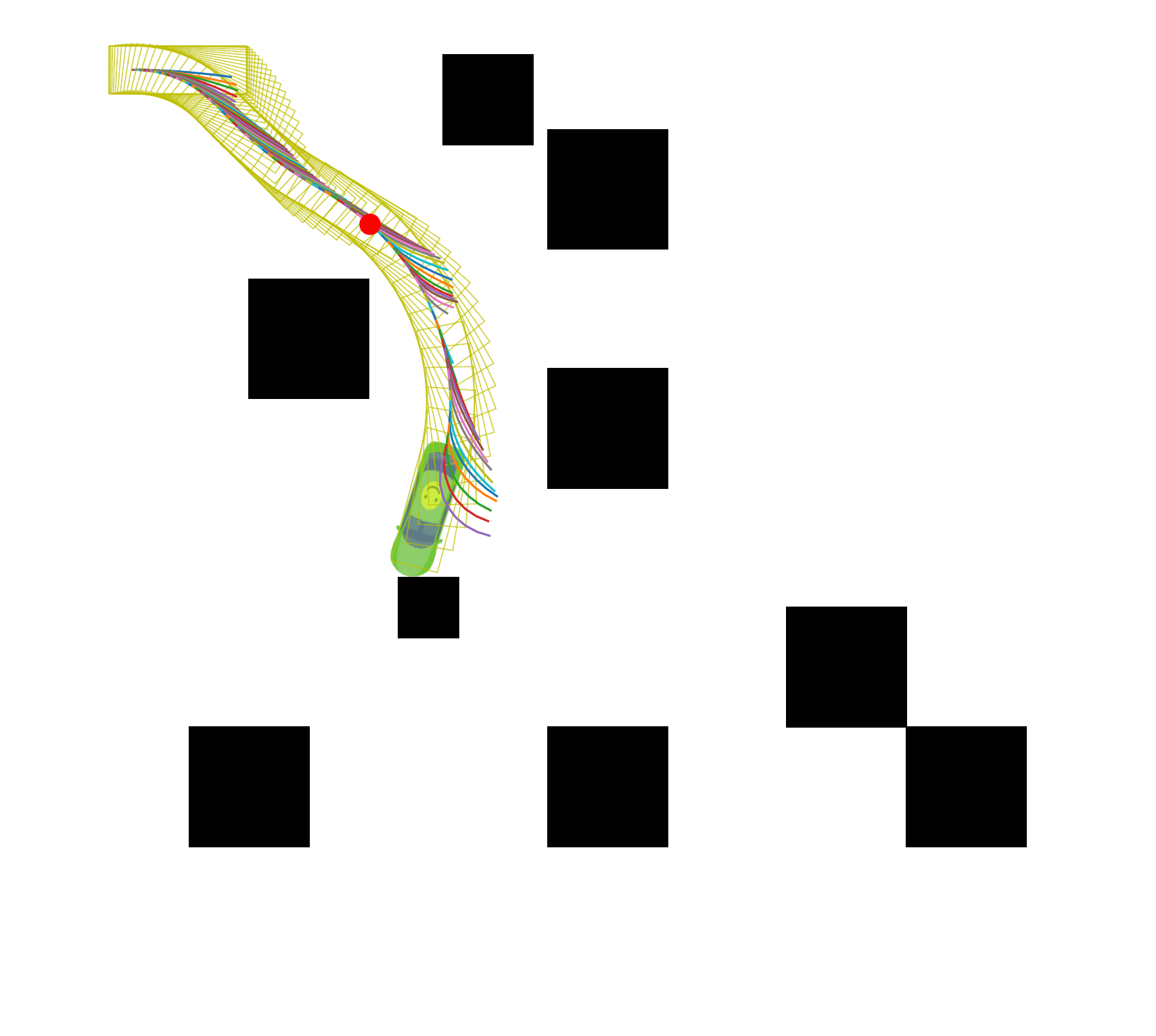}%
		\label{fig4a}}
	% \hfil
	\subfloat[RDA Planner + MDD-I]{\includegraphics[width=1.75in]{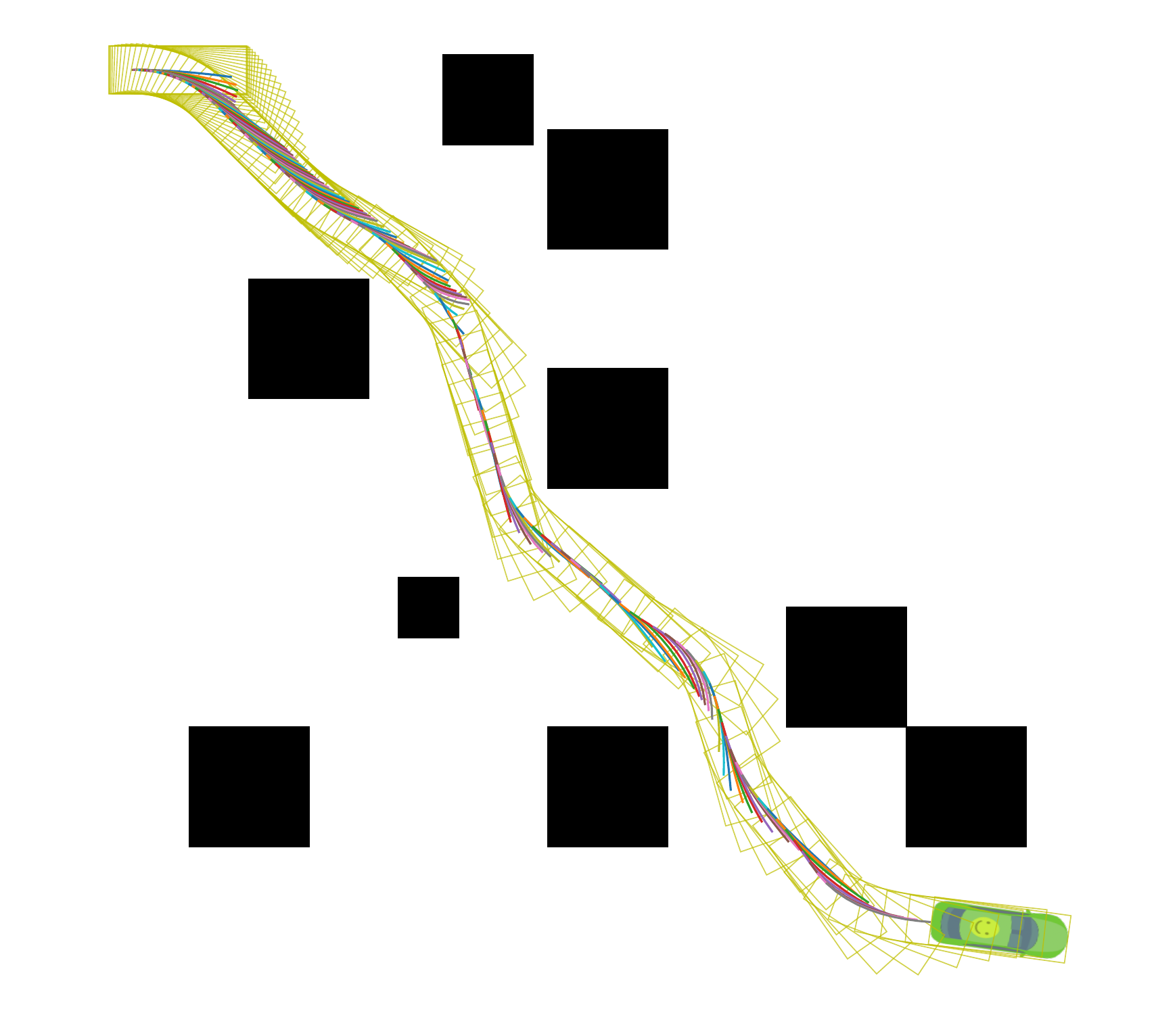}%
		\label{fig4b}}
        % \hfil
	\subfloat[Neural Dubins Model + MDD-I]{\includegraphics[width=1.75in]{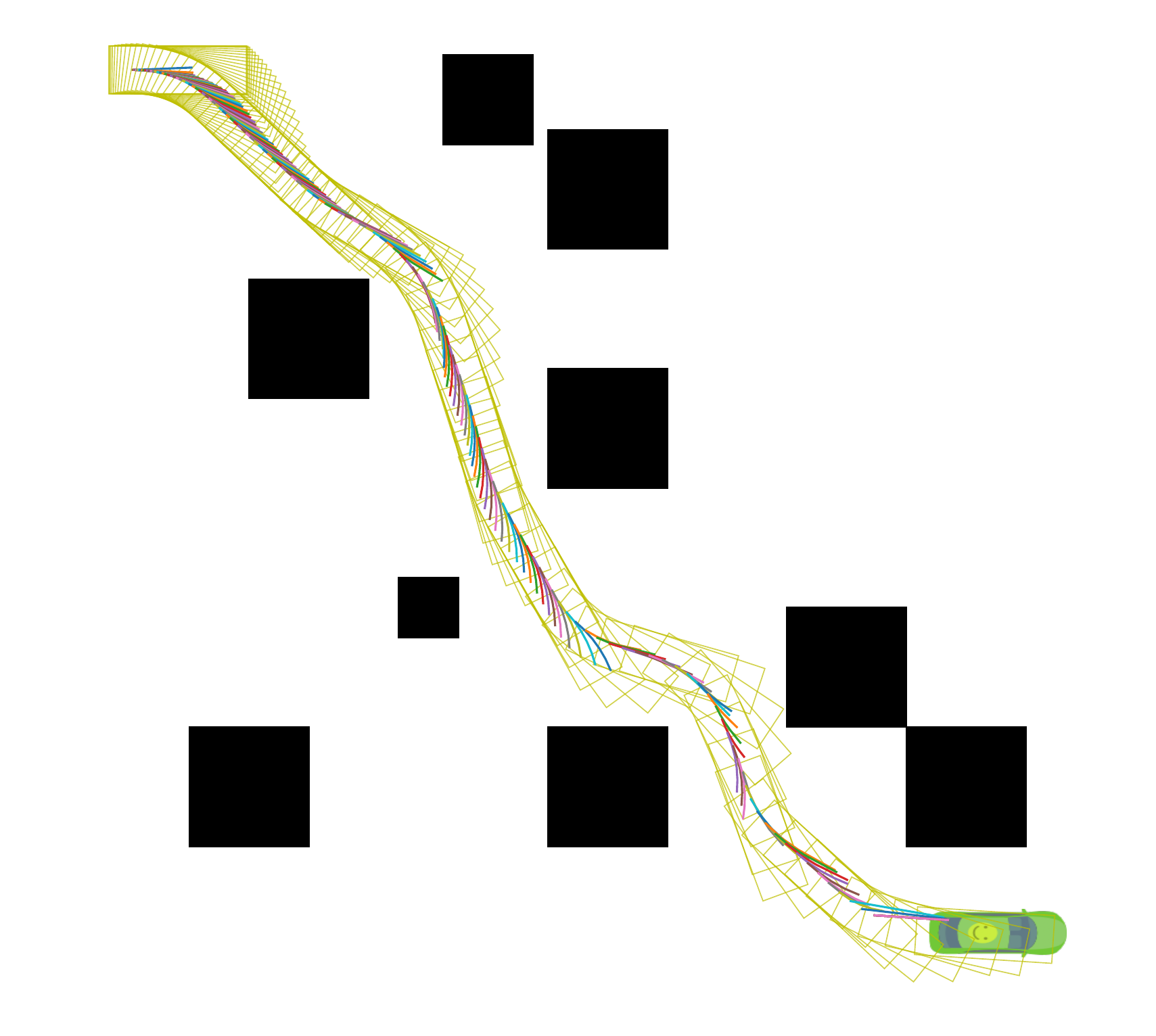}%
		\label{fig4c}}
        % \hfil 
	\subfloat[Neural Dubins Model + MDD-II]{\includegraphics[width=1.75in]{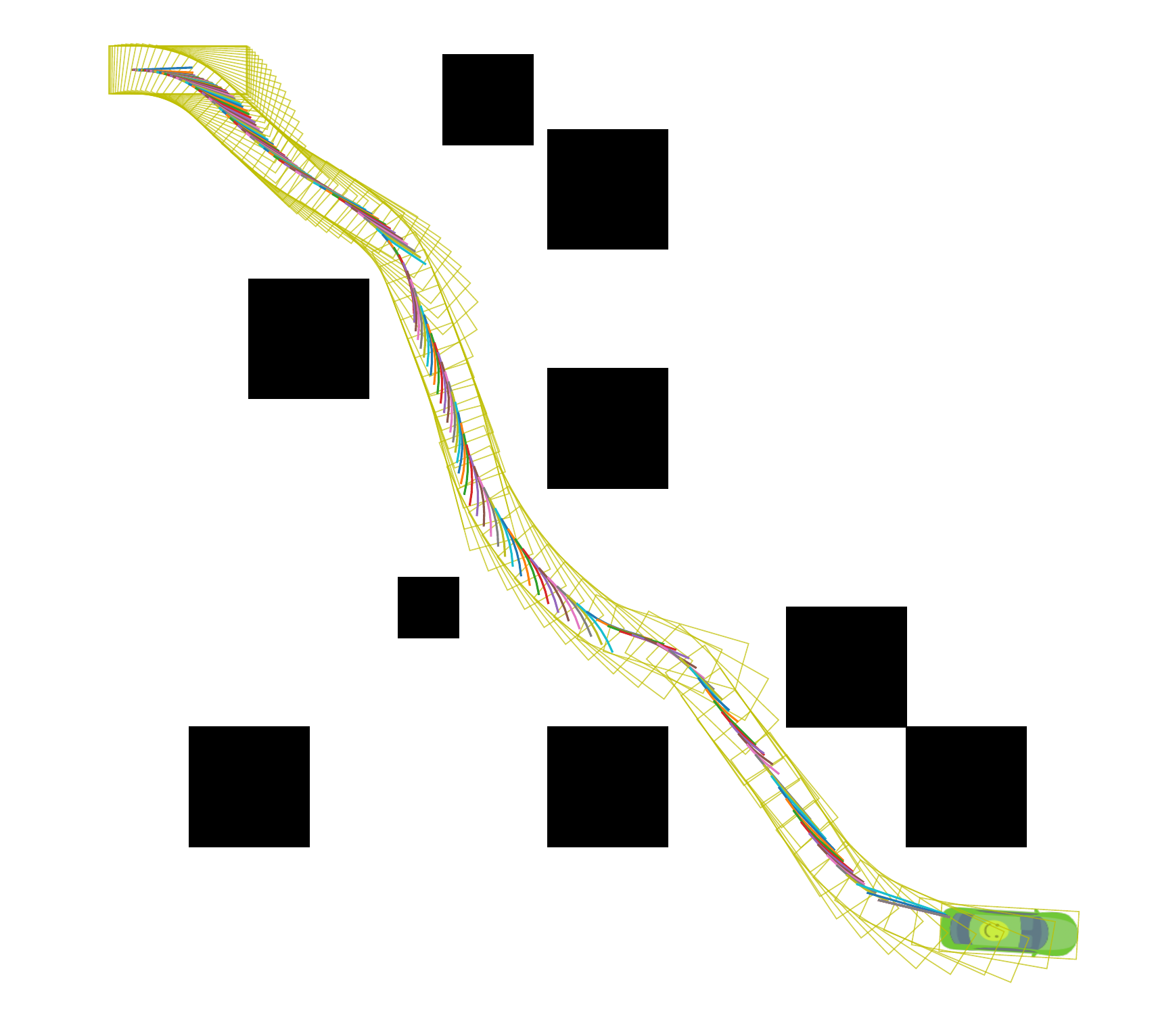}%
		\label{fig4d}} 
        \\
        % \centering
        \subfloat[RDA Planner + MPC-DCBF]{\includegraphics[width=1.6in]{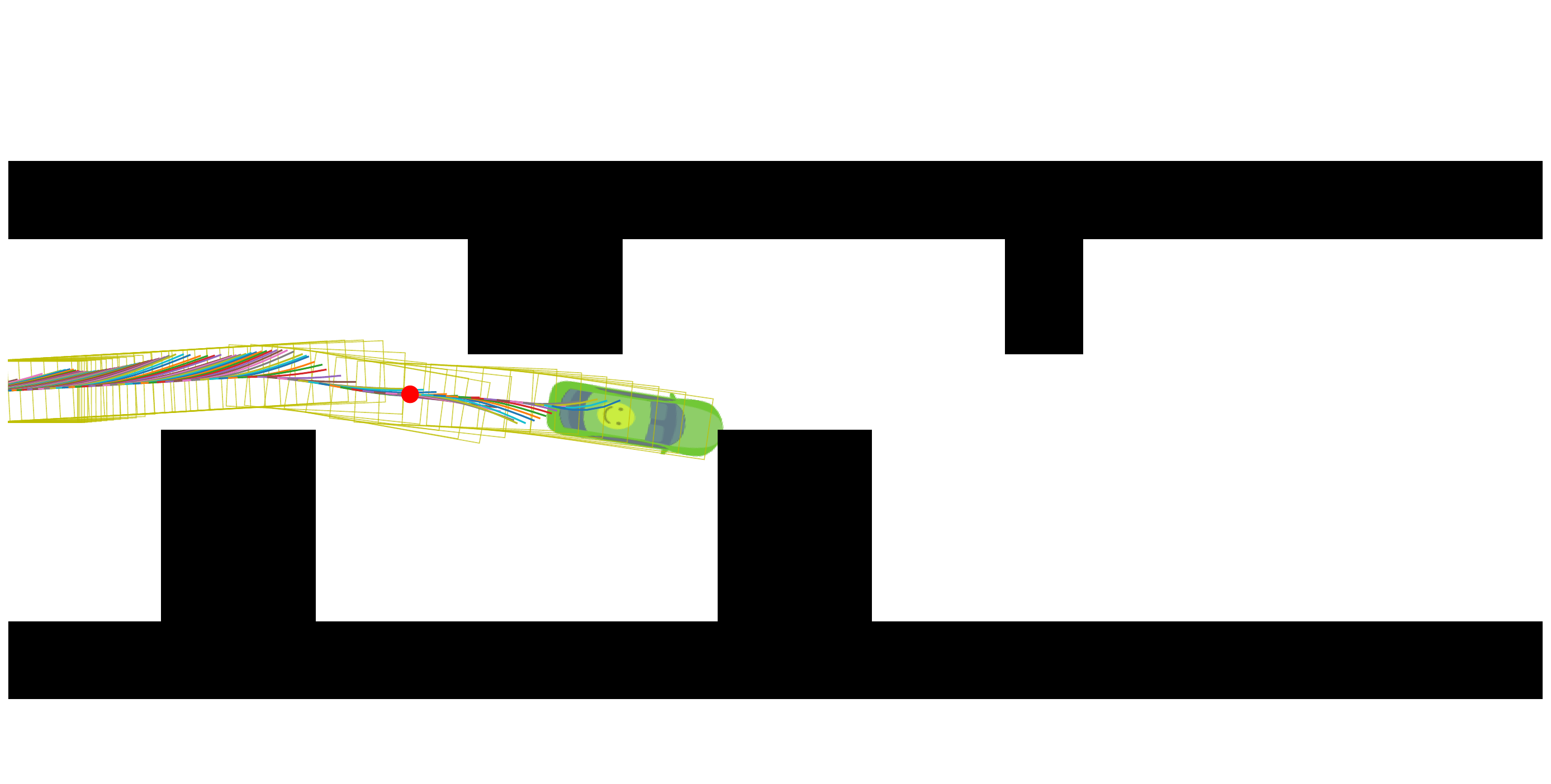}          \label{fig4e}} 
        \hspace{0.4mm}
        % \hfil
        \subfloat[RDA Planner + MDD-I]{\includegraphics[width=1.61in]{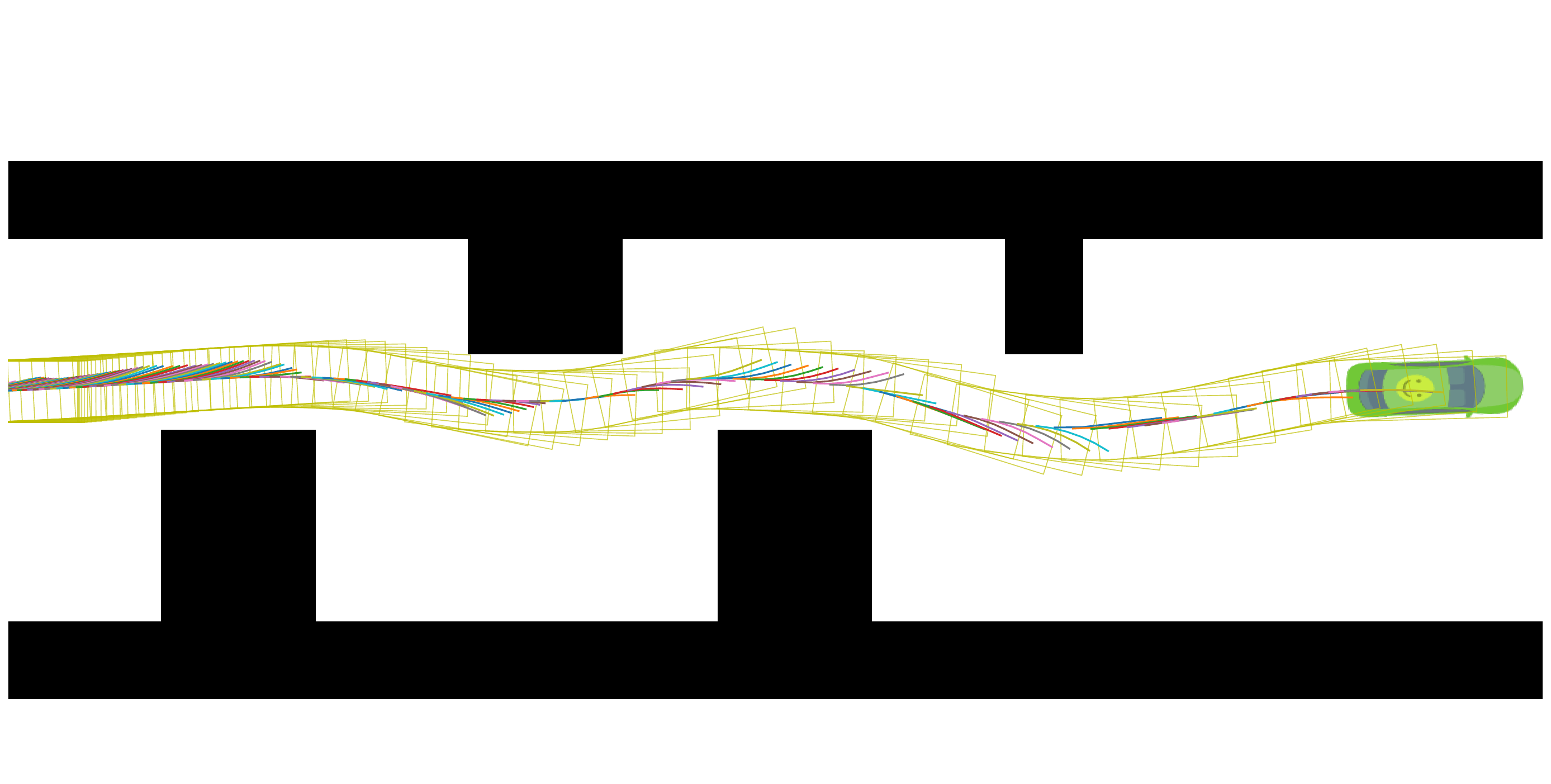}%
		\label{fig4f}}
        \hfil
	\subfloat[Neural Dubins Model + MDD-I]{\includegraphics[width=1.61in]{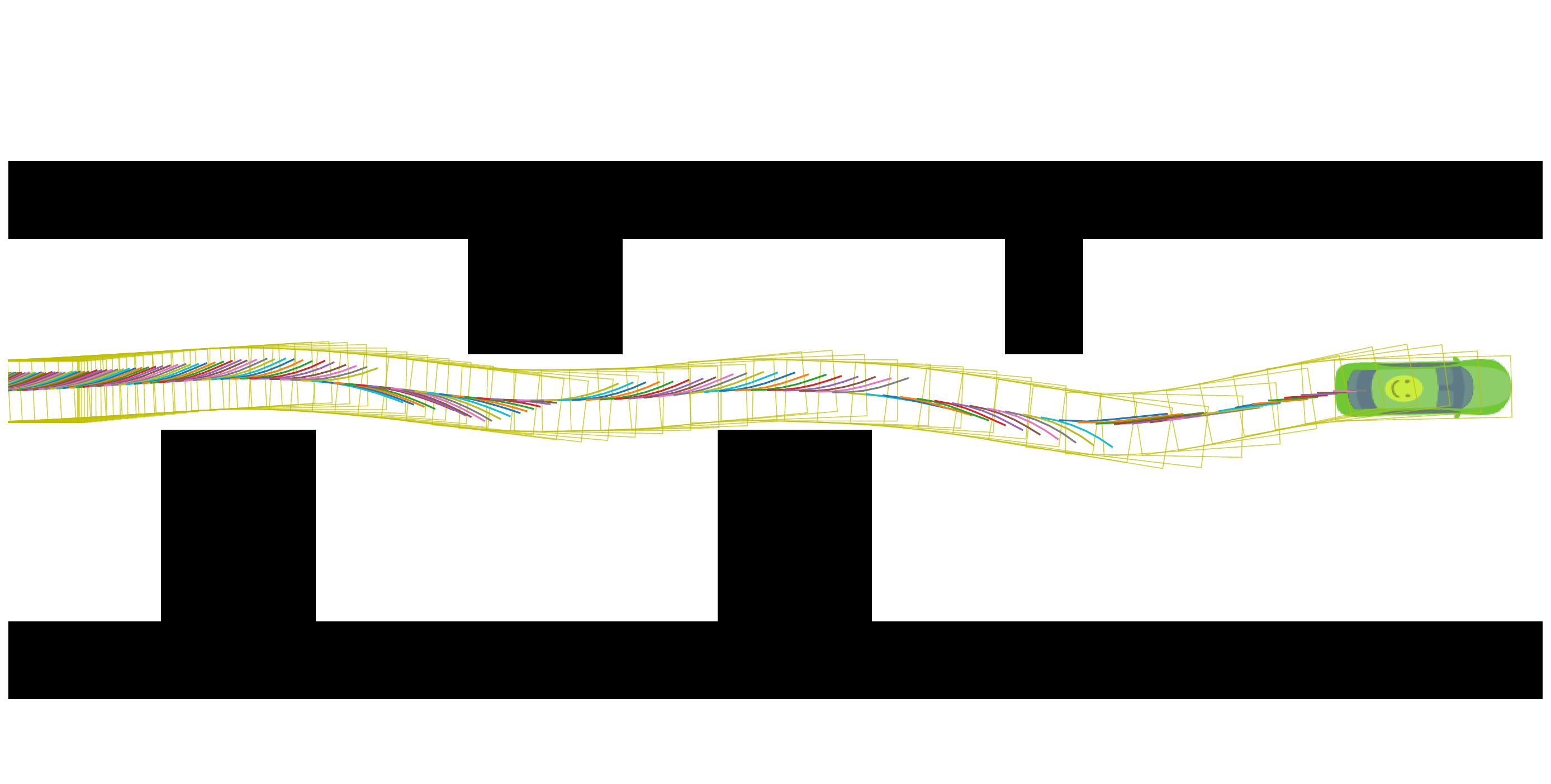}%
		\label{fig4g}}
        \hfil
	\subfloat[Neural Dubins Model + MDD-II]{\includegraphics[width=1.62in]{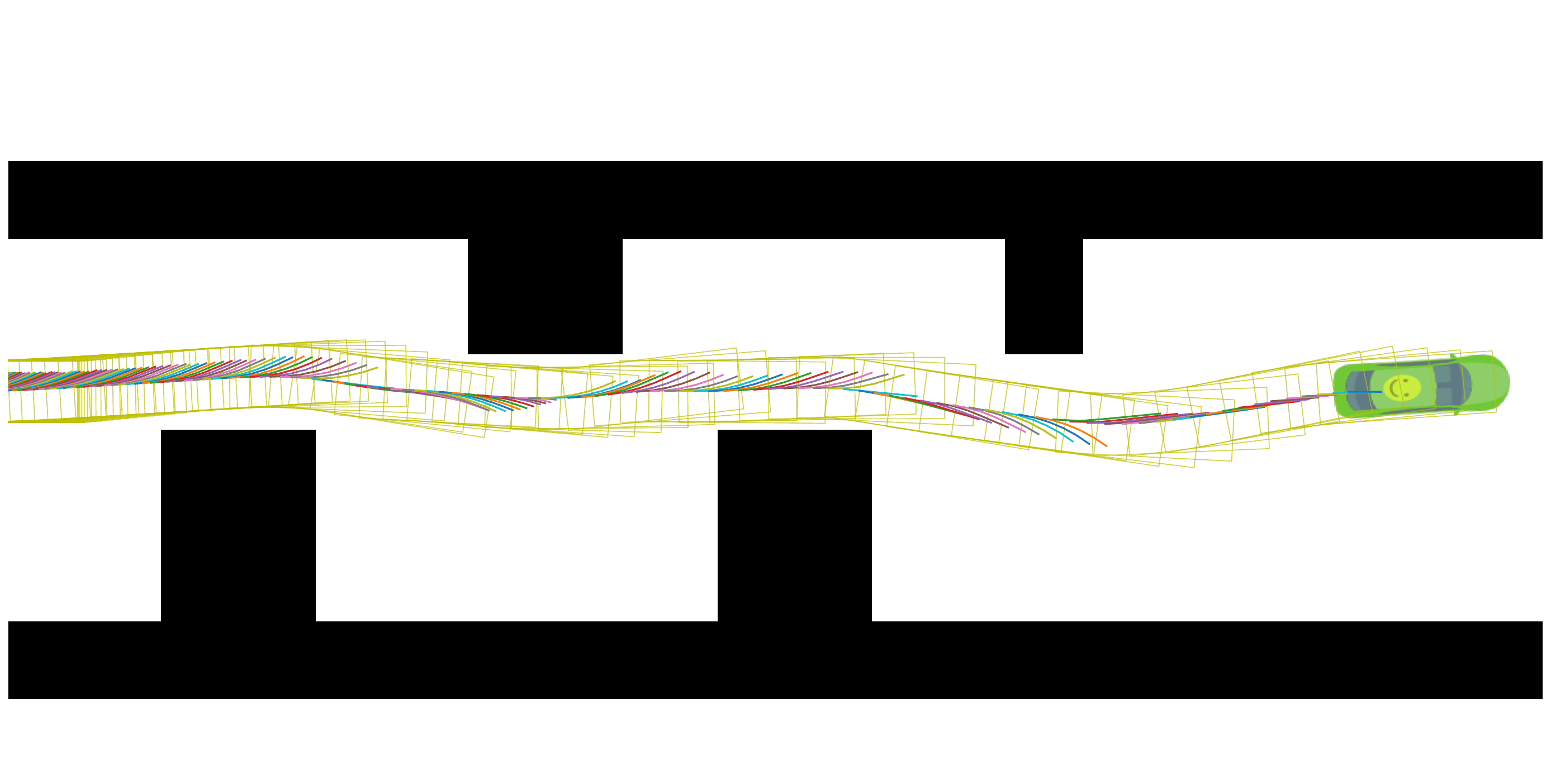}%
		\label{fig4h}}
	\caption{Comparison of trajectories among different algorithms in numerical simulations.}
	\label{fig4}  
\end{figure*}

\subsection{Datasets Construction and Training Details}

Due to the scarcity of high-quality path planning datasets for deep learning in unstructured environments, we have constructed a dataset that includes information on obstacles and the robot's motion trajectories. 
The generation of motion trajectories within the dataset is facilitated by the RDA Planner \cite{han2023rda}.
The simulation environment is configured as a 50 m $\times$ 50 m square workspace. Start and goal poses are stochastically sampled within bounded regions located at the square’s opposite corners. 
The obstacle set comprises four to six obstacles. A subset is uniformly sampled across the entirety of the 50 m $\times$ 50 m workspace, while the remaining obstacles are stochastically positioned within a narrow corridor centered on the arena’s diagonal axis.
Obstacle dimensions are likewise randomized within a bounded interval, ensuring that neither excessive size nor diminutive scale compromises dataset fidelity or task validity.
Subsequently, the planner is employed to generate collision-free trajectories. The obstacle configurations and corresponding trajectory data are then aggregated to constitute a single training instance.

For ease of representation, obstacles are uniformly modeled as circles, with their information consisting of the center and radius of the circles. Trajectory information includes the coordinates of the robot's rear axle center and the robot's heading angle. We have excluded data where collisions occurred during motion, resulting in trajectory lengths shorter than a predetermined threshold.

We have generated a total of 12,000 instances for the training set and 3,000 instances for the test set.

Training was performed on a workstation equipped with an NVIDIA RTX 4060 GPU and 16 GB of system memory. The learning rate was set to 0.001, and the Adam optimizer was adopted to accelerate convergence. A batch size of 32 was employed to balance memory utilization and computational efficiency. Finally, the model was trained for 200 epochs to ensure full convergence.

\subsection{Numerical Simulation}

We have prepared the following four combinations of motion control algorithms for comparative experiments: (1) The RDA is utilized as the planner, with the MPC-DCBF serving as the controller.
(2) The RDA is utilized as the planner, with the MDD-I serving as the controller. (3) The Neural Dubins Model is utilized as the planner, with the MDD-I serving as the controller. (4) The Neural Dubins Model is utilized as the planner, with the MDD-II serving as the controller. We have set the parameters for the aforementioned algorithms to \( N = 11 \), \( N_{{\text{CBF}}} = 10 \), and \( \gamma = 0.9 \). The optimization problems are implemented in Python using CasADi\cite{andersson2019casadi} as the modeling language and solved with IPOPT \cite{biegler2009ipopt} on Ubuntu 18.04. 

When the degree of constraint violation of a numerical solution is less than the feasibility tolerance of the solver, the solver will regard this solution as a feasible point. Consequently, violations of the obstacle avoidance constraint may occur, leading to collisions.

We validated our proposed framework in intelligent robot simulator (ir-sim)\footnote{\url{https://github.com/hanruihua/ir_sim}}, which is a Python-based simulator for robotic algorithm development.
We represent obstacles with black polygons, denote the area traversed by the robot with a yellow border, and illustrate the reference trajectory provided by the planner to the controller with colored lines. Comparative simulations were conducted in both square scenarios and line scenarios to comprehensively evaluate the proposed framework.

\begin{figure}[ht]
\centering
 \subfloat[Square Scenario]{\includegraphics[width=0.22\textwidth]{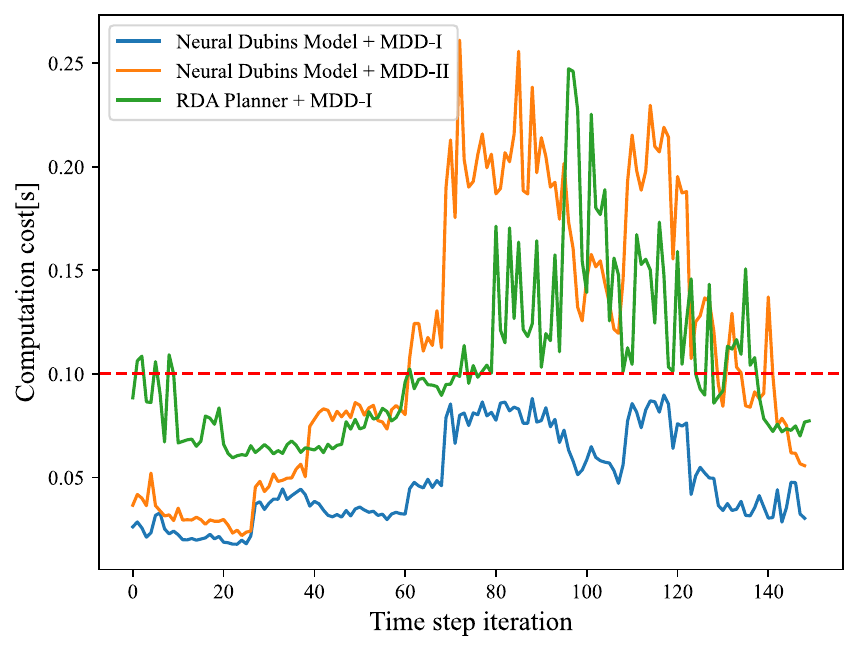}}%
		\label{fig5a}
 \subfloat[Line Scenario]{\includegraphics[width=0.22\textwidth]{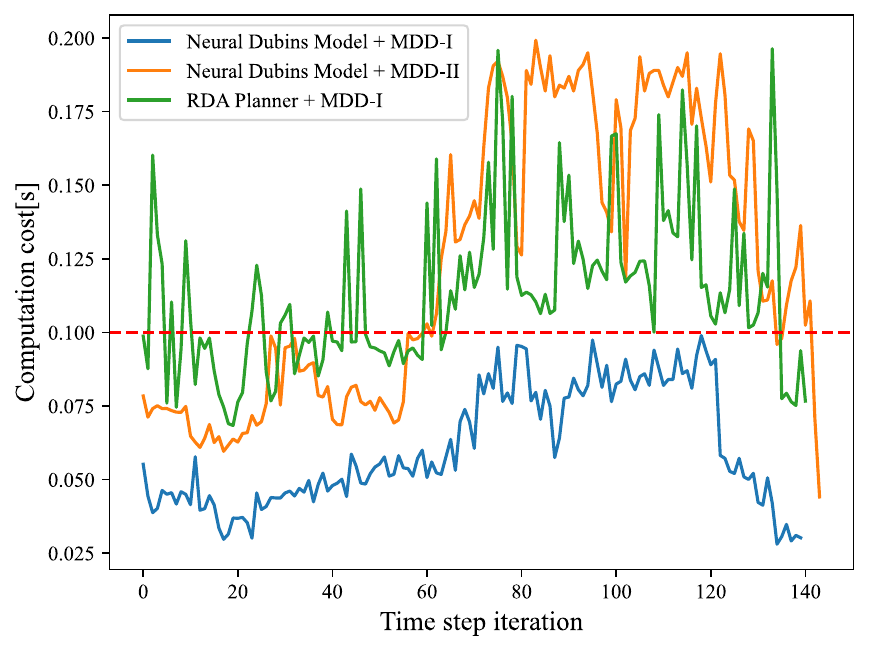}}%
		\label{fig5b}
  \caption{Comparison of computation cost of three algorithms. It illustrates the deviation of the computational cost of three algorithms at each time step from the $0.1$s benchmark line in Fig. \ref{fig4}.}
  \label{fig5}
\end{figure}

\begin{figure*}[!t]
	\centering
	\subfloat[RDA Planner + MDD-I]{\includegraphics[width=1.7in]{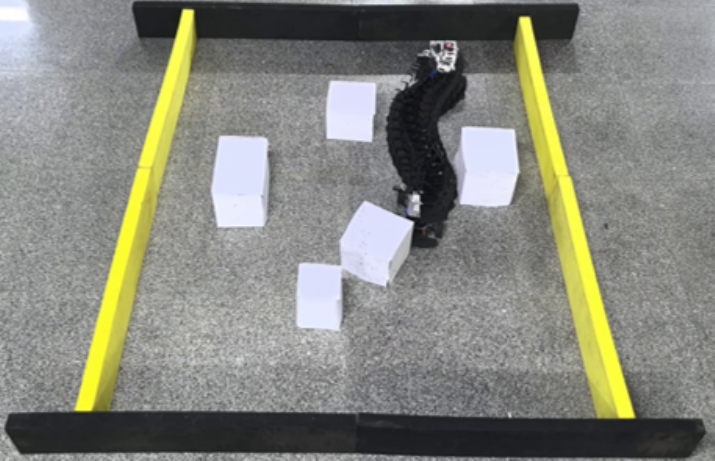}%
		\label{fig6a}}
	\hfil
	\subfloat[Neural Dubins Model + MDD-I]{\includegraphics[width=1.7in]{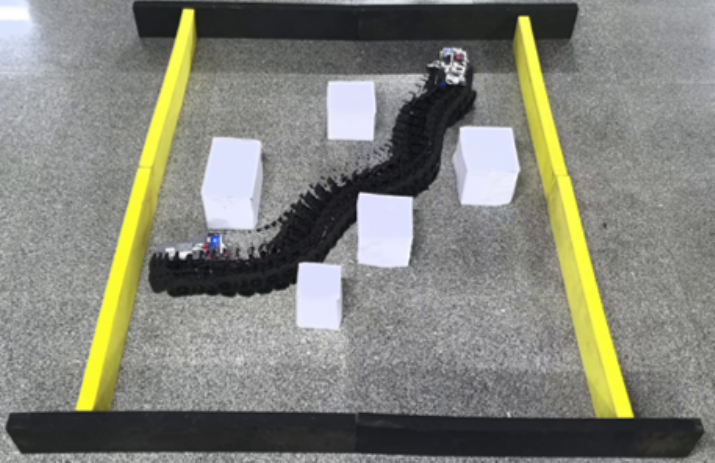}%
		\label{fig6b}}
        \hfil
	\subfloat[Neural Dubins Model + MDD-II]{\includegraphics[width=1.7in]{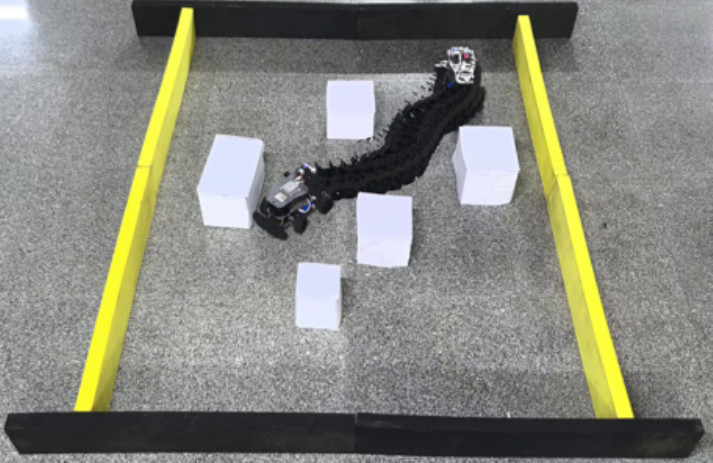}%
		\label{fig6c}}
        \hfil
	\subfloat[Line Scenario with 4 Obstacles]{\includegraphics[width=1.7in]{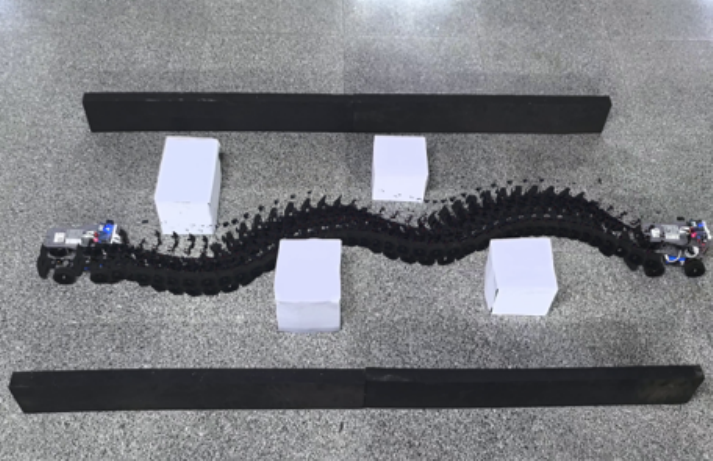}%
		\label{fig6d}} \\
        \subfloat[Square Scenario with 6 Obstacles]{\includegraphics[width=1.7in]{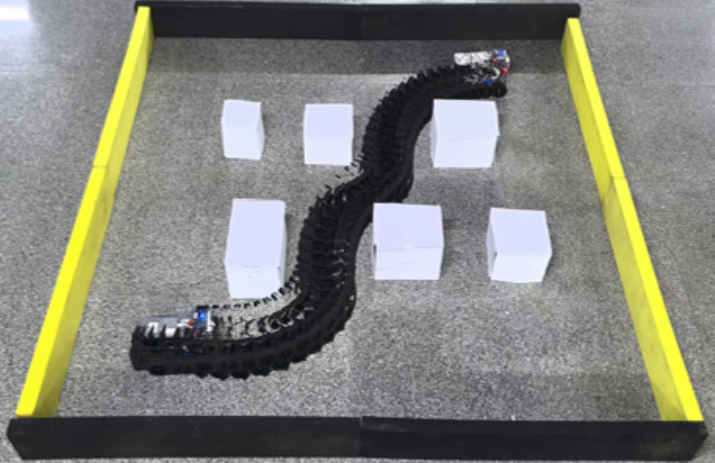}%
		\label{fig6e}}
	\hfil
	\subfloat[Square Scenario with 7 Obstacles]{\includegraphics[width=1.7in]{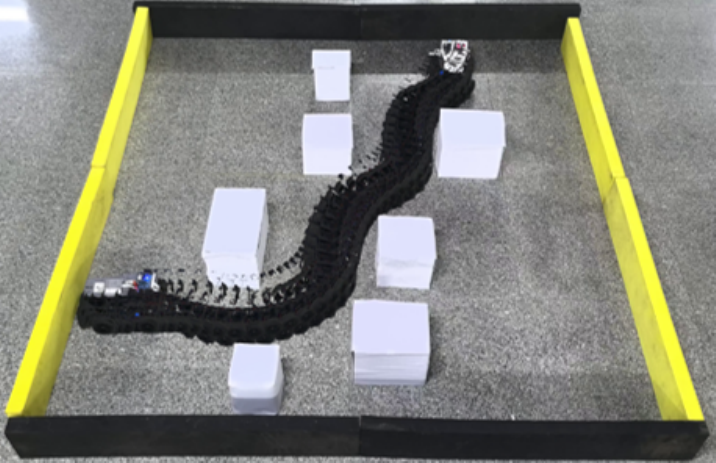}%
		\label{fig6f}}
        \hfil
	\subfloat[Square Scenario with 8 Obstacles]{\includegraphics[width=1.7in]{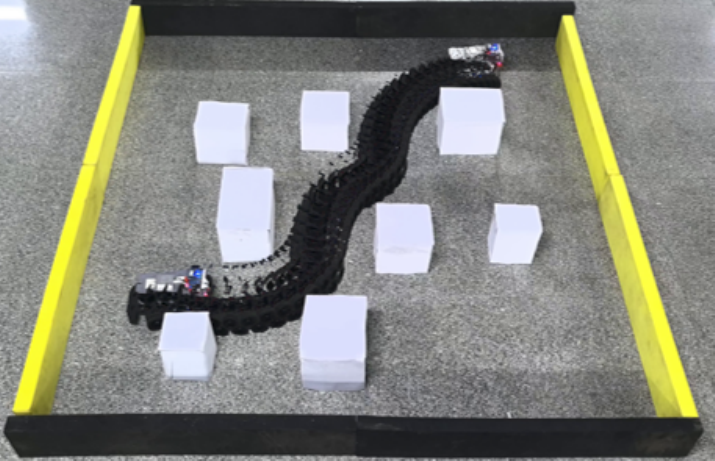}%
		\label{fig6g}}
        \hfil
	\subfloat[Line Scenario with 6 Obstacles]{\includegraphics[width=1.7in]{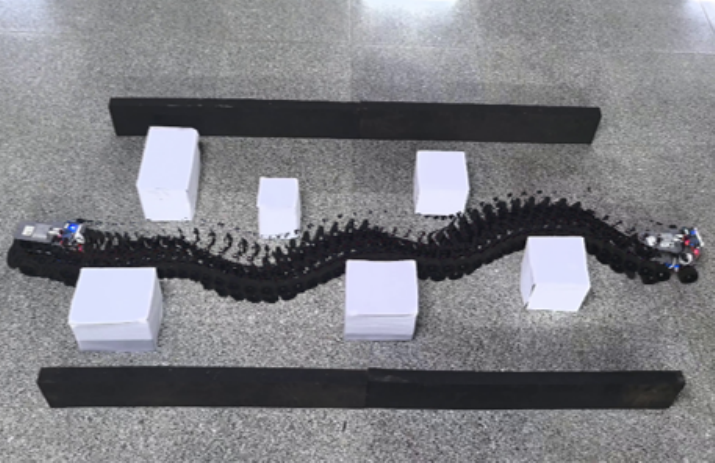}%
		\label{fig6h}} 
        
	\caption{Comparison of trajectories among different algorithms in real-world experiments. During the experiments, the control module frequency was set to 10 Hz.}
	\label{fig6}
\end{figure*}

In Fig. 4\subref{fig4a} and Fig. 4\subref{fig4e}, the red dots denote locations where the solver fails to obtain a feasible solution, demonstrating the limited adaptability of MPC-DCBF in complex scenarios. 
Compared with Fig. 4\subref{fig4b} and Fig. 4\subref{fig4f}, the \mbox{MDD-I} controller demonstrates superior adaptability to complex environments, exhibiting no solver failures. Similarly, inspection of Fig. 4\subref{fig4c} - \subref{fig4d} and Fig. 4\subref{fig4g} - \subref{fig4h} reveals that, upon integrating the Neural Dubins Model, both frameworks consistently generate collision-free trajectories that satisfy all imposed constraints. 
Nevertheless, the three frameworks exhibit markedly disparate computational overheads. In obstacle-dense regions, the average solution times for RDA Planner with MDD-I and Neural Dubins Model with MDD-II cluster around 0.15 seconds, whereas Neural Dubins Model with MDD-I consistently solves within 0.10 seconds. These quantitative discrepancies are further illustrated in Fig. \ref{fig5}.
% Fig. 4\subref{fig4b} shows that MDD-I can adapt to complex environments, however, due to the Dubins curve not accounting for obstacles, the controller collided while following the reference trajectory. Fig. 4\subref{fig4c} - 4\subref{fig4d} illustrate that, upon employing the Neural Dubins Model, all two algorithms are capable of planning a suitable collision-free trajectory. This demonstrates that the Neural Dubins Model provides an appropriate reference path. However, there is a significant difference in computation time among the two algorithms, especially in densely obstacle-populated areas, where the computation time for MDD-II often reaches around 0.15 seconds, whereas the computation time for MDD-I does not exceed 0.1 seconds. Fig. \ref{fig5} reflects the disparity in their computation times.

To quantitatively and visually assess the disparities in success rates and average solution times among the algorithms, we conducted comparative experiments across 50 randomly generated square scenarios and 50 randomly generated line scenarios. The aggregated results are presented in Table \ref{table1} and Table \ref{table2}.

Here, the success rate refers to the ratio of the number of scenarios where the robot successfully reaches the destination without collision to the total number of scenarios. 
Across all successful scenarios, we separately report the single-step average runtime for both the planner and the controller, and additionally computed the mean of the maximum single-step execution times. 
% The average solving time is defined as the mean value of the solution times across all successful scenarios. The average maximum solving time refers to the average of the maximum single-step solving time across all successful scenarios.
\begin{table}[ht]
\caption{Performance across 50 random square scenarios}
\label{table1}
\begin{center}
\begin{tabular}{c|c|c|c|c}
\hline
    & Success Rate & \makecell[c]{Planner} & \makecell[c]{Controller} & \makecell[c]{Max} \\
\hline
\makecell[c]{RDA \\ MPC-DCBF}  & 0.16 & - & - & -\\ 
\hline
\makecell[c]{RDA \\ MDD-I} & 0.88 & 54ms & 32ms & 167ms \\
\hline
\makecell[c]{Neural Dubins \\ MDD-I}  & 0.86 & 5ms & 33ms & 86ms\\
\hline
\makecell[c]{Neural Dubins \\ MDD-II}  & 0.88 & 5ms & 84ms & 181ms\\
\hline
\end{tabular}
\end{center}
\end{table}

\begin{table}[ht]
\caption{Performance across 50 random line scenarios}
\label{table2}
\begin{center}
\begin{tabular}{c|c|c|c|c}
\hline
    & Success Rate & \makecell[c]{Planner} & \makecell[c]{Controller} & \makecell[c]{Max} \\
\hline
\makecell[c]{RDA \\ MPC-DCBF}  & 0.12 & - & - & - \\
\hline
\makecell[c]{RDA \\ MDD-I} & 0.76 & 63ms & 44ms & 176ms\\
\hline
\makecell[c]{Neural Dubins \\ MDD-I}  & 0.76 & 5ms & 48ms & 92ms \\
\hline
\makecell[c]{Neural Dubins \\ MDD-II}  & 0.80 & 5ms & 93ms & 193ms\\
\hline
\end{tabular}
\end{center}
\end{table}

Owing to the prohibitively low success rate of RDA Planner with MPC-DCBF, its runtime statistics lack statistical reliability and are therefore excluded from the reported results.
The remaining three algorithms exhibit comparable success rates. However, the Neural Dubins Model markedly outperforms the RDA Planner in computational efficiency. Moreover, MDD-I yields a significantly lower single-step solution time than MDD-II. Collectively, the Neural Dubins Model paired with MDD-I demonstrates a pronounced advantage in overall solution speed.

\subsection{Real-World Experiments}

\begin{figure}[ht]
\centering
 \includegraphics[width=0.4\textwidth]{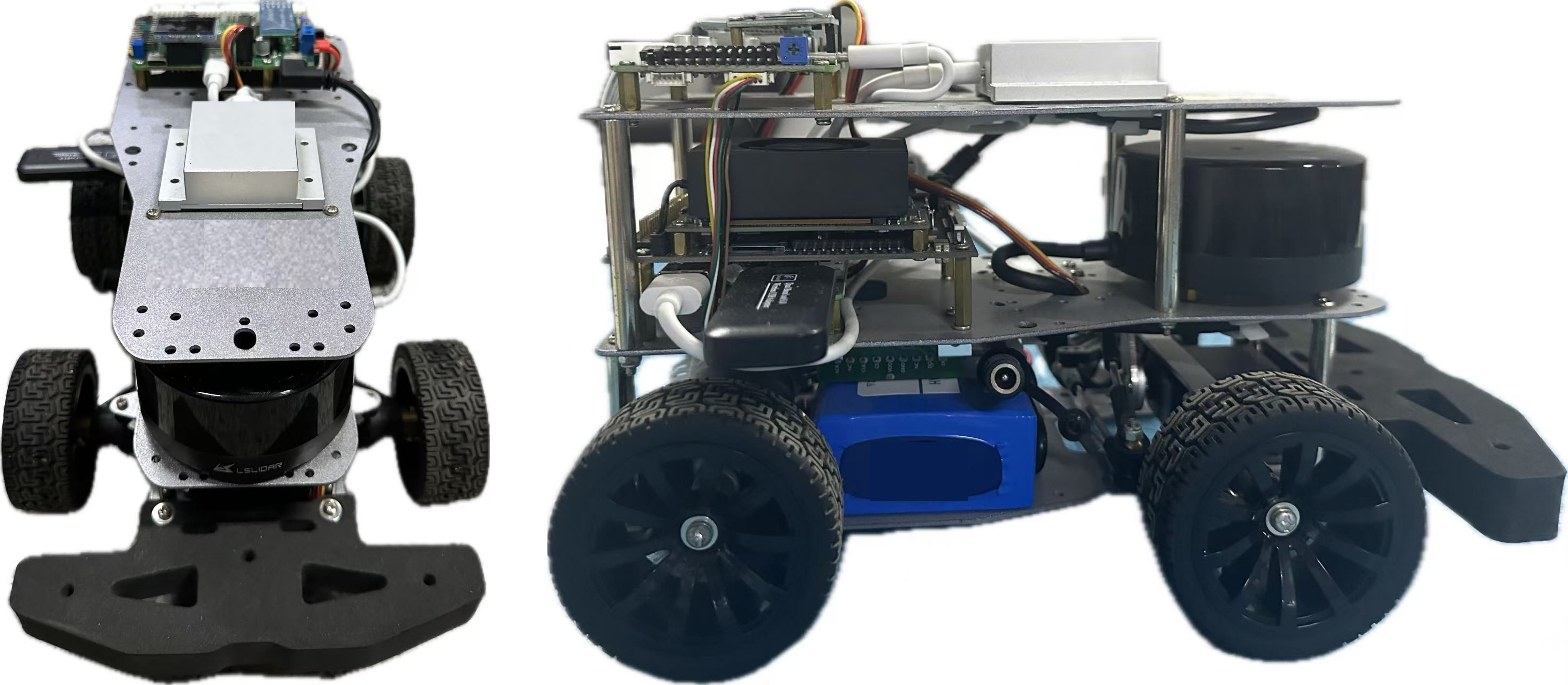}
  \caption{The Ackermann mobile robot.}
  \label{fig7}
\end{figure}

To validate the practical performance of the NMPCB framework in real-world scenarios, experiments were conducted on the robot depicted in Fig. \ref{fig7}. This robot is equipped with four wheels and employs Ackermann steering. Its sensor suite comprises a 2D LiDAR, an IMU, and wheel odometry. We utilized an NVIDIA Jetson Nano as the computing platform.

Utilizing the aforementioned hardware, we carried out real-world experimental validation.
Inspection of \mbox{Fig. 6\subref{fig6a} - \subref{fig6c}} reveals that, because the control module’s execution frequency exceeds the algorithm’s solution frequency, neither RDA Planner with MDD-I nor Neural Dubins Model with MDD-II is able to reach the goal. In contrast, Neural Dubins Model with MDD-I demonstrates superior real-time performance and successfully reaches the goal.
\mbox{Fig. 6\subref{fig6d} - \subref{fig6h}} show that the Neural Dubins Model with MDD-I was further evaluated under a diverse set of challenging scenarios. In every trial the robot successfully reached the goal, thereby confirming the algorithm’s robust feasibility in complex environments.
% In Fig. 6\subref{fig6a}, the MPC-DCBF algorithm faced persistent solution failures in the presence of multiple obstacles, ultimately resulting in a collision.
% In Fig. 6\subref{fig6b}, although no solution failures occurred, collisions still took place due to the use of the Dubins curve as the reference trajectory.
% Comparing Fig. 6\subref{fig6c} and Fig. 6\subref{fig6d}, the MDD-II algorithm resulted in a collision due to the robot's control module frequency exceeding the solution frequency of MDD-II. However, under the same control module frequency, the MDD-I algorithm demonstrated superior real-time performance, successfully reaching the goal.

\section{Conclusion}

In this paper, we propose NMPCB, a motion control framework that integrates neural networks with optimal control. This framework serves dual purposes as both a local planner and a controller. While ensuring the safety of the robot's motion, it also demonstrates commendable real-time performance compared to other methods. For future research, we will optimize the Neural Dubins Model to enhance its capabilities in path planning. Additionally, we will focus on constructing optimization formulations of different forms to accelerate the solution process.

\bibliographystyle{IEEEtran}
\bibliography{IEEEabrv,reference}

\end{document}